\documentclass[sigconf]{acmart}
\usepackage{diagbox}
\usepackage{array}
\usepackage{tablefootnote}
\usepackage{multirow}
\usepackage{multicol}
\usepackage{tikz}
\newcommand*\circled[1]{\tikz[baseline=(char.base)]{
            \node[shape=circle,draw,inner sep=0.5pt] (char) {\small#1};}}

\newcommand{\nop}[1]{}

\acmConference[Conference acronym 'XX]{Make sure to enter the correct
  conference title from your rights confirmation emai}{June 03--05,
  2018}{Woodstock, NY}
\acmPrice{15.00}
\acmISBN{978-1-4503-XXXX-X/18/06}

\begin{document}

\title{What do LLMs Know about Financial Markets\nop{the Stock Market}? A Case Study on Reddit Market Sentiment Analysis}

\author{Xiang Deng}
\authornote{Work done while interning at Google.}
\email{deng.595@buckeyemail.osu.edu}
\affiliation{%
  \institution{The Ohio State University}
  \state{OH}
  \country{USA}
}

\author{Vasilisa Bashlovkina}
\email{vasilisa@google.com}
\affiliation{%
  \institution{Google Research}
  \state{NY}
  \country{USA}}

\author{Feng Han}
\email{bladehan@google.com}
\affiliation{%
  \institution{Google Research}
  \state{NY}
  \country{USA}}

\author{Simon Baumgartner}
\email{simonba@google.com}
\affiliation{%
  \institution{Google Research}
  \state{NY}
  \country{USA}}

\author{Michael Bendersky}
\email{bemike@google.com}
\affiliation{%
  \institution{Google Research}
  \state{CA}
  \country{USA}}


\begin{abstract}
Market sentiment analysis on social media content requires knowledge of both financial markets and social media jargon, which makes it a challenging task for human raters. The resulting lack of high-quality labeled data stands in the way of conventional supervised learning methods. Instead, we approach this problem using semi-supervised learning with a large language model (LLM). Our pipeline generates weak financial sentiment labels for Reddit posts with an LLM and then uses that data to train a small model that can be served in production. We find that prompting the LLM to produce Chain-of-Thought summaries and forcing it through several reasoning paths helps generate more stable and accurate labels, while using a regression loss further improves distillation quality. With only a handful of prompts, the final model performs on par with existing supervised models. Though production applications of our model are limited by ethical considerations, the model's competitive performance points to the great potential of using LLMs for tasks that otherwise require skill-intensive annotation.
\end{abstract}

\ccsdesc[500]{Computing methodologies~Natural language processing}
\ccsdesc[500]{Information systems~Social networks}

\keywords{Sentiment Analysis, Social Media, Finance, Large Language Model, Natural Language Processing}

\maketitle

\section{Introduction}
\begin{figure*}
    \centering
    \vspace{-1em}
    \includegraphics[width=\linewidth]{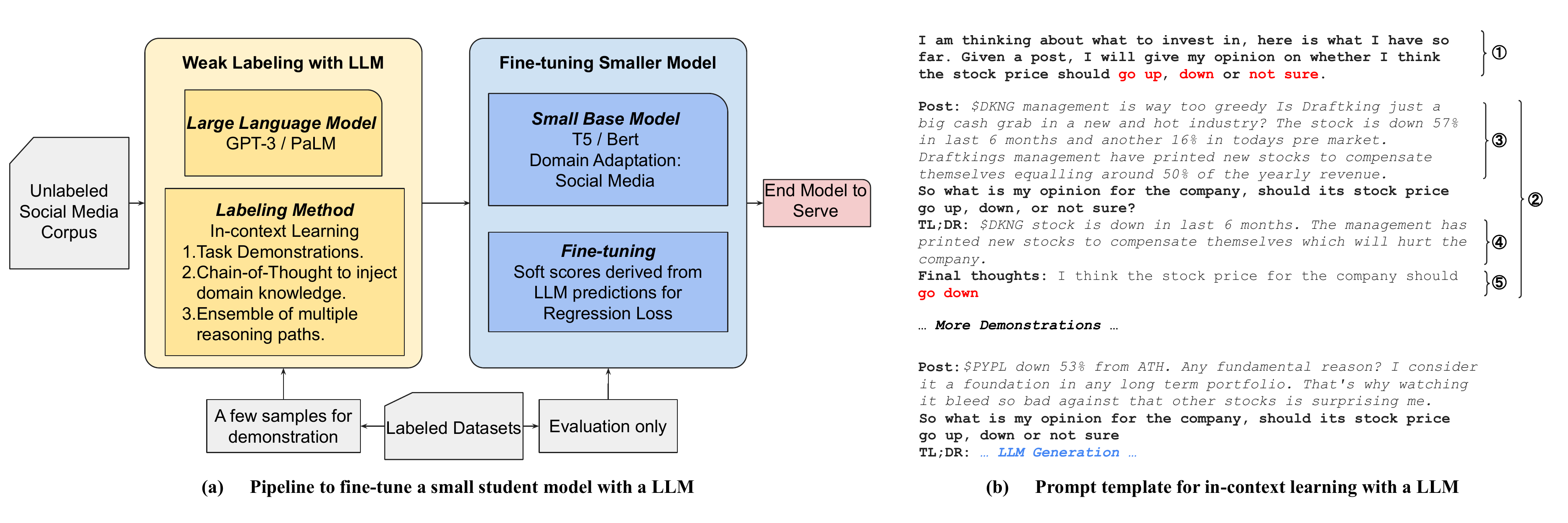}
    \vspace{-3em}
    \caption{Our overall pipeline (a), and prompt design for in-context learning (b).}
    \vspace{-1em}
    \label{fig:prompt}
\end{figure*}
Social media platforms such as Reddit and Twitter contain insights about financial markets, for example in the form of posts that express financial expectations for a particular company. We define the financial sentiment of a post about a company as positive (bullish) if the author of the post
has a favorable outlook for the company, negative (bearish) if their outlook for the company is negative, and neutral otherwise.
Financial (market) sentiment analysis aims to automatically extract the financial performance expectations conveyed in the text.

One particular challenge for market sentiment analysis on social media is the lack of high-quality labeled data, which arises because the annotation requires both finance domain knowledge and an understanding of social media jargon. Previous study has found that "bearish" or "bullish" tags selected by the authors of the posts themselves are often not accurate, and even finance experts may hold different opinions during annotation~\cite{chen-etal-2020-issues}. Our in-house annotation effort revealed the same issue, with raters only agreeing with each other around 70\% of the time.

In the meantime, large language models (LLMs) such as GPT-3~\cite{gpt3} and PaLM~\cite{palm} gained popularity in recent years for their impressive aptitude for in-context learning~\cite{gpt3, emergent}. An LLM can perform textual tasks with just a few examples demonstrating what needs to be done, often achieving results similar to those of state-of-the-art supervised models in a wide range of applications. 

Inspired by this development, we investigate the use of LLMs to bootstrap a market sentiment analysis model for social media content with minimal human annotation efforts. We select Reddit as the target social media platform, which, unlike other platforms used in existing works (Twitter and Stocktwits), has a broader coverage of topics, ranging from market events to analysis and user investing actions, with both short user comments and long "due diligence" posts. To annotate Reddit posts with weak financial sentiment labels, we use LLM in-context learning~\cite{gpt3, palm} with Chain-of-Thought~\cite{COT} reasoning and repeated generation~\cite{Wang2022SelfConsistencyIC} for more stable predictions. Since the LLM is too large and slow to be used in a production setting, we distill it into a smaller student model. We find that if we aggregate multiple predictions for a single example into a soft score and use a regression loss, we can make use of more data and get a smoother precision-recall curve. Even though we are able to serve our student model online and control its precision by setting the prediction threshold, the model's application is limited by ethical considerations stemming from the high-stakes nature of investment decisions. Nevertheless, compared with models fine-tuned on existing market sentiment datasets, our model trained with only weakly-labeled Reddit data not only improves on the challenging in-domain testing data from Reddit, but also generalizes well across datasets and performs on par with the supervised counterparts.
\section{Related Work}
\noindent\textbf{Market Sentiment Analysis.}
Existing work on the task of extracting opinions about financial entities from text roughly falls into two categories: lexicon-based methods that associate individual words with sentiment labels~\cite{loughran2011liability, chen2018ntusd, loughran2016textual}, and machine learning methods that train a supervised model with labeled data~\cite{liu2021finbert, yang2020finbert, araci2019finbert}. However, they either show unsatisfactory performance or demand huge amounts of labeled data that is hard to acquire in practice. In this work, we conduct an exploratory study of leveraging the in-context learning ability of LLMs to overcome the data challenge.

\noindent\textbf{In-context Learning with LLM.}
In-context learning, or prompting, refers to the capability of LLMs to perform tasks by making predictions conditioned on a few input-output examples without updating any model parameters~\cite{gpt3}. Many recent works have studied the underlying mechanism of in-context learning~\cite{emergent, min2022rethinking, Xieincontext}, and how to improve the few-shot performance~\cite{COT, zhou2022least, Li2022OnTA, Wang2022SelfConsistencyIC}. We adapt some of these techniques for market sentiment analysis and design a pipeline that puts them into practice.
\section{Methodology}
\subsection{In-context Learning with LLM}
Though our target task is analyzing the author's overall financial outlook for a company, in our target domain, social media, users normally discuss financial performance in terms of stock price movement. We use the domain-adapted proxy task of extracting stock price movement expectations in our LLM prompt, as shown in Figure~\ref{fig:prompt}:

\noindent\textbf{Task description} \circled{1}, which simulates the task setting and familiarizes the LLM with the target domain.

\noindent\textbf{Demonstrations} \circled{2}, which illustrate the task via multiple input-output examples (\circled{3}, \circled{5}). The output \circled{5} is verbalized to make it similar to the examples seen by the model during pre-training and then converted to actual categorical labels during post-processing.

Our preliminary study shows that while this prompt design can already yield reasonable results, the prediction is unstable and sensitive to the exact wording of the prompt. In particular, we notice that the results vary a lot if we simply shuffle the order of demonstrations, which means that the model struggles to truly understand the user's opinion~\cite{min2022rethinking}. To counter this instability, we incorporate \textbf{Chain-of-Thought} (COT)~\cite{COT} reasoning into our setting. COT was originally designed to improve the multi-step reasoning ability of LLMs by explicitly instructing the model to generate intermediate reasoning steps. 
While we do not need multi-step reasoning for market sentiment analysis, we use COT to make the LLM summarize the author's finance-related arguments TL;DR-style (\circled{4}), thus implicitly forcing it to recall relevant financial domain knowledge before drawing a conclusion (\circled{5}).
Because users often cite multiple, sometimes conflicting arguments in their posts, we use temperature sampling~\cite{ackley1985learning, ficler-goldberg-2017-controlling} instead of greedy decoding during generation and repeat it multiple times to produce varying reasoning paths, giving the model a chance to focus on different lines of argument. As a result, each example gets assigned multiple, potentially inconsistent labels~\cite{Wang2022SelfConsistencyIC}. We use majority voting to get the final prediction for in-context learning, and describe how to better leverage the multiple predictions for distillation in Section~\ref{sec:distill}.

\subsection{Bootstrapping a Market Sentiment Model with LLM}
\label{sec:distill}
While in-context learning with LLMs has shown impressive results during offline evaluation ~\cite{gpt3}, it is impractical to serve such large models in production. A commonly used compression method is to first generate a large weakly-labeled dataset using the larger teacher model and then train a smaller student model in a supervised fashion~\cite{gou2021knowledge}. We do notice that in some cases there are complex or ambiguous posts where the LLM assigns the weak label incorrectly. It is also the case that for those hard examples, the LLM makes inconsistent predictions when exploring different reasoning paths. A straightforward way to leverage this pattern would be to filter out weakly-labeled examples that have any inconsistency among the labels assigned via different LLM reasoning paths. However, such filtering may cost us many potentially useful examples and cause the student model to overfit to the remaining easy cases. Instead, we view the agreement ratio between the multiple labels of a single example as a soft score of sentiment polarity and train the student model to predict this score with a regression loss. 
\section{Experimental Results}
\subsection{Experimental Setup}
\label{sec:exp_setup}
\noindent\textbf{Problem Formulation.}
We study market sentiment analysis as a three-way classification task. The market sentiment of a post about a particular company is defined as positive (bullish) if the author's outlook for it is favorable, negative (bearish) if their outlook is negative, and neutral otherwise.

\noindent\textbf{Datasets.}
\begin{table}[t]
    \centering
    \caption{Data Statistics.}
    \label{tab:data}
    \vspace{-0.5em}
    \resizebox{0.9\columnwidth}{!}{
    \begin{tabular}{lccc}
    \toprule
         &  FiQA News & FiQA Post & Reddit Testing\\
    \midrule
    \# Total & 370 & 674 & 100\\
    \% Neg / Neu / Pos  & 34/-/66 & 35/-/65 & 39/42/19\\
    Avg. Length  & 9.7 & 13.4 & 83.0\\
    \bottomrule
    \end{tabular}}
    \vspace{-1em}
\end{table}
For both distillation and evaluation, we use Reddit posts labeled as finance-related by a proprietary topic classifier. We filter posts based on the popularity of the mentioned stock in an internal system and randomly sample 20,000 posts for distillation. Since there are no existing datasets for market sentiment analysis on Reddit, we sample another 100 posts for evaluation, which are annotated by three in-house experts who have both knowledge of investing terms and experience with Reddit.

We also experiment with the widely used FiQA benchmark~\cite{Maia2018WWW18OC}, which contains two subtasks: FiQA-News with news headlines, and FiQA-Post with microblogs from Twitter and Stocktwits. We convert it to a binary classification task with the original sentiment scores.\footnote{Examples with sentiment scores greater than 0 are considered positive, and negative otherwise. We remove the few examples with a sentiment score of 0 and those mentioning multiple stocks. This drops 66 examples for News and 1 example for Post.} Since the original testing set is private, we split the original training set into training, validation, and testing following an 80/10/10 ratio.

Statistics for all the datasets are summarized in Table~\ref{tab:data}.

\noindent\textbf{Baselines.}
Our backbone model is Charformer~\cite{tay2021charformer} (CF), a character-level T5~\cite{raffel2020exploring}. We further pre-train CF on social media content.
We consider the following baselines: (i) our backbone model fine-tuned on FiQA, (ii) PaLM in-context learning with COT and majority-vote aggregation over 8 reasoning paths, and (iii) two widely used existing market sentiment models: FinBERT-HKUST~\cite{yang2020finbert} and FinBERT-ProsusAI~\cite{araci2019finbert}\footnote{We use the models released on Huggingface ModelHub: \\\url{https://huggingface.co/ProsusAI/finbert}, \\\url{https://huggingface.co/yiyanghkust/finbert-tone}}.

\noindent\textbf{Implementation Details.}
We use PaLM-540B~\cite{palm} as the LLM for in-context learning and weak labeling. We randomly select 6 examples as demonstrations and remove them from the test set when evaluating PaLM. For Reddit, we select two examples for each sentiment category and manually write the COT reasoning. For FiQA, we select three examples for each category, and use the "Aspect Snippet" in the original dataset as COT reasoning. For each input, we run the generation 8 times with a temperature of 0.5 to produce different reasoning paths and predictions. For distillation, we keep weakly-labeled examples for which the LLM makes 5 or more consistent predictions, and fine-tune the Charformer model on 17K Reddit posts labeled with soft scores aggregated from the 8 PaLM predictions for each example. We use a learning rate of 1e-4 and a batch size of 64. We apply a regression head to the final CF encoder layer and drop the decoder. The final CF model in Table~\ref{tab:main} has 102M parameters while both FinBERT models have 110M parameters. Ablations on the number reasoning paths, filtering, and fine-tuning objectives can be found in Figure~\ref{fig:COT},~\ref{fig:Regression} and Table~\ref{tab:ap}.

\subsection{Results}
\label{sec:results}
\begin{table}[t]
    \centering
    \caption{Accuracy on benchmark datasets, see Section~\ref{sec:exp_setup} for more details.}
    \label{tab:main}
    \vspace{-0.5em}
    \resizebox{0.8\columnwidth}{!}{
    \begin{tabular}{lccc}
    \toprule
         &  FiQA News & FiQA Post & Reddit\\
    \midrule
    CF - FiQA News & 75.7 & 69.1 & 42.0\\
    CF - FiQA Post & 86.5 & 85.3 & 40.0\\
    \midrule
    FinBERT-ProsusAI~\cite{araci2019finbert} & 81.1 & 73.5 & 48.0\\
    FinBERT-HKUST~\cite{yang2020finbert} & 75.7 & 67.6& 50.0 \\
    \midrule
    PaLM $COT \times 8$ & 97.3 & 95.6 &72.0 \\
    CF - Distilled PaLM & 83.8& 77.9& 69.0\\
    \bottomrule
    \end{tabular}}
    \vspace{0em}
\end{table}
\begin{figure}
    \centering
    \includegraphics[width=0.8\linewidth]{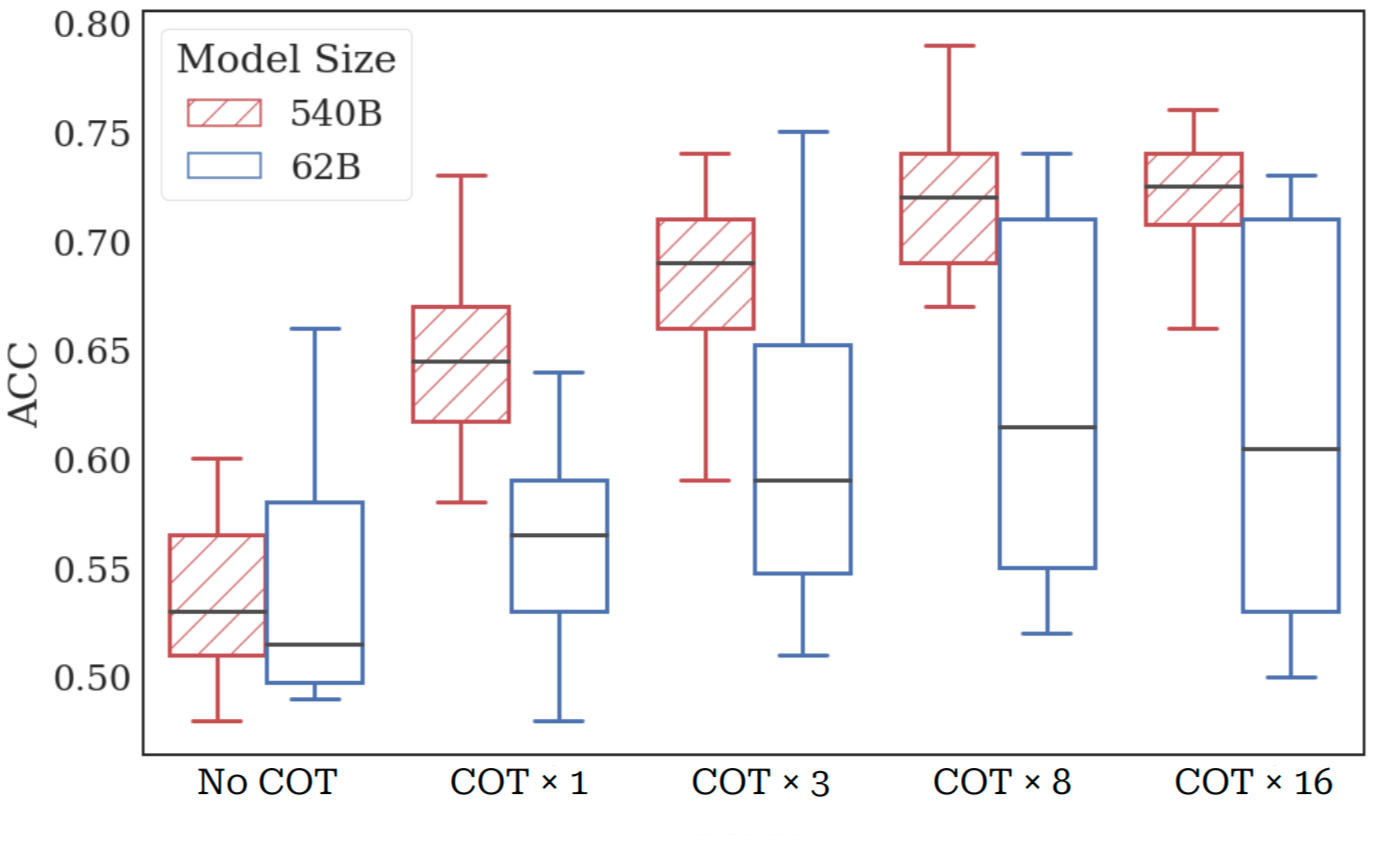}
    \vspace{-1em}
    \caption{Ablation for In-context Learning. Here we compare LLMs of different sizes, effect of COT, and the use of multiple reasoning paths. {\small{$COT \times N$}} stands for generating $N$ reasoning paths with majority voting to aggregate the predictions.}
    \label{fig:COT}
    \vspace{-1.5em}
\end{figure}
\noindent\textbf{Overall.}
Table~\ref{tab:main} summarizes the main results. Here we report accuracy on all datasets. First, we can see that the Reddit dataset is more challenging than the FiQA datasets, likely due to longer posts and multifaceted user opinions. The PaLM model performs very well on all datasets with only 6 demonstration examples. Our student model fine-tuned on weakly labeled Reddit data is able to effectively transfer the knowledge from the LLM and outperform all supervised baselines on the Reddit dataset. At the same time, our model generalizes well to the FiQA dataset despite only being fine-tuned on Reddit posts. In summary, the experiments show promising results of leveraging LLMs for market sentiment analysis. With only a handful of labeled examples for demonstration, we can bootstrap a small student model that performs on par with or better than existing state-of-the-art models of servable size.

\noindent\textbf{Ablation on In-context Learning.}
\begin{table}[t]
    \centering
    \caption{Average precision for Positive and Negative labels. Here we compare using classification (CLS) and regression (RGR) loss at different intra-label agreement thresholds.}
    \label{tab:ap}
    \vspace{-0.5em}
    \resizebox{0.7\columnwidth}{!}{
    \begin{tabular}{llcccc}
    \toprule
    \multicolumn{2}{l}{Agreement}& 8 & 7 & 6 & 5\\
    \midrule
    \multicolumn{2}{l}{\# Examples}&
    6,240&10,474&14,152&17,456\\
    \midrule
    \multirow{2}{*}{Pos}
    &CLS  &80.5 & 75.8 & 71.4 & 76.9 \\
    &RGR  &74.2 &78.5 &81.7&84.2 \\
    \midrule
    \multirow{2}{*}{Neg}
    &CLS  &68.0 &64.0& 47.4 &57.9 \\
    &RGR  &54.3 &61.8 &61.7&65.5 \\
    \bottomrule
    \end{tabular}}
    \vspace{0em}
\end{table}
Figure~\ref{fig:COT} demonstrates the importance of using chain-of-thought (COT) reasoning and repeating the generation for in-context learning. Here we use the same demonstration examples but shuffle their order to get different prompts. First, we see that in-context learning is sensitive to the prompt design: even with only the order of examples changed, the final performance varies a lot. 
Second, using COT to have PaLM summarize the post's main arguments as a TL;DR greatly improves the performance.
Asking the LLM to generate multiple reasoning paths and aggregating the predictions further boosts the performance, as it allows the model to explore different aspects of the user's opinion. Finally, model size influences the effectiveness of COT reasoning. While the 62B and 540B PaLM models have similar performance with the base prompt, the 540B model benefits much more from COT, likely because its superior generational ability allows it to produce more useful intermediate thinking steps.

\begin{figure}
    \centering
    \includegraphics[width=0.9\linewidth]{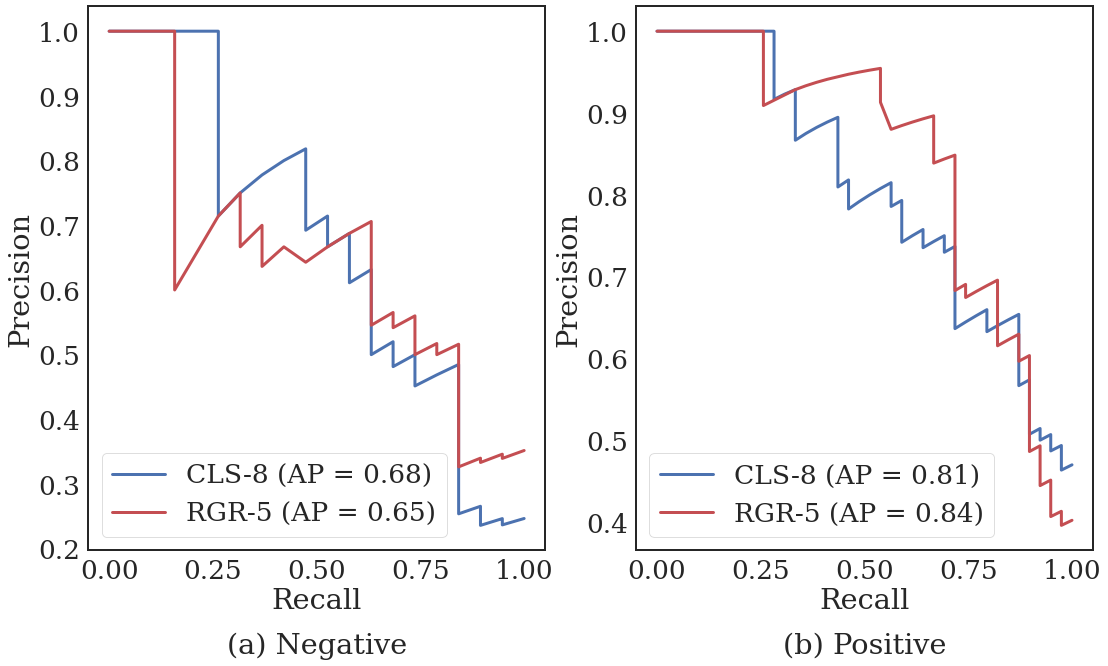}
    \vspace{-1em}
    \caption{Precision-Recall curve. Here we compare models that achieve the best average precision (as shown in Table~\ref{tab:ap}) using Classification (CLS-8) and Regression (RGR-5) loss.}
    \label{fig:Regression}
    \vspace{-1.5em}
\end{figure}
\noindent\textbf{Ablation on Distillation Methods.}
 We compare filtering the PaLM-labeled data with different intra-label agreement thresholds. As we can see from Table~\ref{tab:ap}, exposing the student model to examples with inconsistent labels hurts its performance even though it gets to see more training data that way. We don't include a full ablation on the student model backbone but we experiment with its loss function. Figure~\ref{fig:Regression} shows that using a regression loss instead of classification is advantageous for two reasons: it better leverages the soft scores from the examples with inconsistent labels and it produces a slightly smoother precision-recall curve. The latter is important for production applications because the smoother curve allows us to pick an operating point with the desired precision.

\subsection{Error Analysis}
\begin{table}[t]
    \centering
    \caption{Confusion Matrix for CF - Distilled PaLM on the Reddit dataset.}
    \label{tab:confusion}
    \vspace{-0.5em}
    \resizebox{0.7\columnwidth}{!}{
    \begin{tabular}{|ll|ccc|}
    \hline
    &&\multicolumn{3}{c|}{Predicted}\\
    && Negative & Neutral & Positive\\
    \hline
    \multirow{3}{*}{\rotatebox[origin=c]{90}{Actual}}
    &Negative  & 9 & 10 & 0 \\
    &Neutral  & 3 & 32& 7\\
    &Positive  & 2& 9&28 \\
    \hline
    \end{tabular}}
\end{table}
We conduct error analysis over the Reddit testing set for our final model (CF - Distilled PaLM). As shown in the confusion matrix in Table~\ref{tab:confusion}, the majority of errors are between neutral and the other two labels, which is less severe than positive/negative errors. We notice that the model struggles when the input contains contradictory arguments or discusses advanced investing actions. Better handling such complicated posts and dynamically incorporating relevant finance domain knowledge could be a subject of future work.

\section{Ethical Considerations}
Applying our model to social media content can make its wealth of financial information more accessible to users. For example, bullish/bearish tags for individual posts can help novice investors orient themselves in the language of r/wallstreetbets.
However, the model's output should not be used to make investment decisions due to the associated risks.
First, the model predicts the wrong sentiment more than 30\% of the time. Second, even if the model doesn't make a mistake, the social media posts it is applied to may convey sentiment that prompts the user to make bad financial decisions. Prior research has found that investors are susceptible to social media advice~\cite{undue_influence} even though the sentiment it carries is a poor predictor of stock prices~\cite{place_your_bets}. Finally, aggregating financial sentiment from social media may amplify malicious behavior like market manipulation. In fact, our model detected a negative sentiment spike for Pfizer in March 2021 when there seemed to be a coordinated effort to promote a rumor that Pfizer shares were getting delisted from NYSE\footnote{\url{https://factcheck.afp.com/doc.afp.com.328D4BT}}. These risks need to be thoroughly addressed and mitigated to ensure that the likely benefits from deploying our model substantially outweigh the foreseeable downsides.

\section{Conclusion}
In this work, we tackle the task of financial sentiment analysis on Reddit with an LLM distilled into a production-friendly student model. With minimal human-annotated data, our classifier performs on par with existing supervised models and generalizes well across other datasets.
The application of our model does pose a product challenge: 
how can we incorporate the model's output responsibly, delivering value to users without misleading them or inadvertently amplifying malicious behavior?
Nevertheless, our investigation highlights the promise of in-context learning with LLMs for textual tasks that are hard for human raters to annotate. Can human raters, instead of simply labeling the data, help design a domain-knowledge-injected prompt teaching the LLM to perform the task, or otherwise "collaborate" with the LLM? How can automatic prompt-tuning further optimize the human-engineered prompt? Exploring the answers to these questions would be a compelling direction for future work.


\onecolumn
\begin{multicols}{2}
\bibliographystyle{ACM-Reference-Format}
\bibliography{ref}
\end{multicols}

\end{document}